\documentclass{article}
\usepackage{spconf,amsmath,graphicx}

\usepackage{enumitem}
\setlist{nosep, leftmargin=14pt}

\usepackage{mwe} 
\usepackage{graphicx}
\usepackage{amsmath}
\usepackage{amsfonts}
\usepackage{multirow}
\usepackage{booktabs} 
\usepackage{bbm}
\usepackage{url}
\usepackage{hyperref}
\usepackage{adjustbox}
\usepackage{color}

\title{Unified Ultrasound Intelligence Toward an End-to-End Agentic System}

\name{Chen Ma, Yunshu Li, Junhu Fu, Shuyu Liang,  Yuanyuan Wang$^{*}$, Yi Guo$^{*}$\thanks{* Corresponding author: yywang@fudan.edu.cn; guoyi@fudan.edu.cn } }
\address{College of Biomedical Engineering, Fudan University, Shanghai, China}

\begin{document}
%
\frenchspacing
\maketitle
\begin{abstract}
Clinical ultrasound analysis demands models that generalize across heterogeneous organs, views, and devices, while supporting interpretable workflow-level analysis. Existing methods often rely on task-wise adaptation, and joint learning may be unstable due to cross-task interference, making it hard to deliver workflow-level outputs in practice. To address these challenges, we present USTri, a tri-stage ultrasound intelligence pipeline for unified multi-organ, multi-task analysis. Stage I trains a universal generalist USGen on different domains to learn broad, transferable priors that are robust to device and protocol variability. To better handle domain shifts and reach task-aligned performance while preserving ultrasound shared knowledge, Stage II builds USpec by keeping USGen frozen and finetuning dataset-specific heads. Stage III introduces USAgent, which mimics clinician workflows by orchestrating USpec specialists for multi-step inference and deterministic structured reports. On the FMC\_UIA validation set, our model achieves the best overall performance across 4 task types and 27 datasets, outperforming state-of-the-art methods. Moreover, qualitative results show that USAgent produces clinically structured reports with high accuracy and interpretability. Our study suggests a scalable path to ultrasound intelligence that generalizes across heterogeneous ultrasound tasks and supports consistent end-to-end clinical workflows. The code is publicly available at: https://github.com/MacDunno/USTri.
\end{abstract}

\section{Introduction}
\label{sec:intro}
Ultrasound is widely used in routine screening and point-of-care diagnosis, but building scalable learning-based ultrasound systems remains difficult in practice~\cite{liu2019deep}. Clinical ultrasound data are highly heterogeneous across organs, views, devices, and acquisition protocols, while downstream objectives span dense delineation, anatomical localization, quantitative measurement, and diagnostic categorization~\cite{brattain2018machine}. This diversity makes it difficult to maintain numerous task-specific models in real-world deployment, and joint training over heterogeneous supervision signals may be unstable and suffer from cross-task interference without careful design~\cite{crawshaw2020multi, standley2020tasks}.

Foundation models have substantially improved transferability in ultrasound imaging~\cite{kirillov2023segment, ma2024segment}. Recent ultrasound foundation models, notably the USFM series~\cite{usfm, tinyusfm}, show strong versatility across organs and task types. However, practical deployment still typically relies on downstream adaptation, and these models alone do not yet constitute a unified, end-to-end pipeline that can reliably support heterogeneous tasks across datasets~\cite{zhang2024challenges}. Moreover, clinical deployment often calls for workflow-level, multi-task capabilities rather than isolated single-pass predictions for one task. Real systems~\cite{goodell2025large} are expected to route requests to appropriate modules, compose multi-step analyses, and return interpretable results. This need motivates agentic inference paradigms that interleave reasoning with tool use~\cite{schick2023toolformer}, especially when paired with biomedical vision-language models (VLM)~\cite{li2023llava}.

To bridge the gap between current ultrasound modeling and real clinical deployment, we present USTri, a tri-stage pipeline tailored to multi-organ, multi-view ultrasound with operator-dependent acquisition and artifact-induced variability. USTri integrates a shared ultrasound representation with efficient specialization, and supports workflow-level quantification via VLM-guided routing to produce structured, interpretable clinical outputs across heterogeneous tasks and datasets.

\section{METHOD}
\label{sec:method}

\subsection{Overview: Tri-Stage Ultrasound Intelligence}

\begin{figure*}[t]
    \centering
    \includegraphics[width=0.9\linewidth]{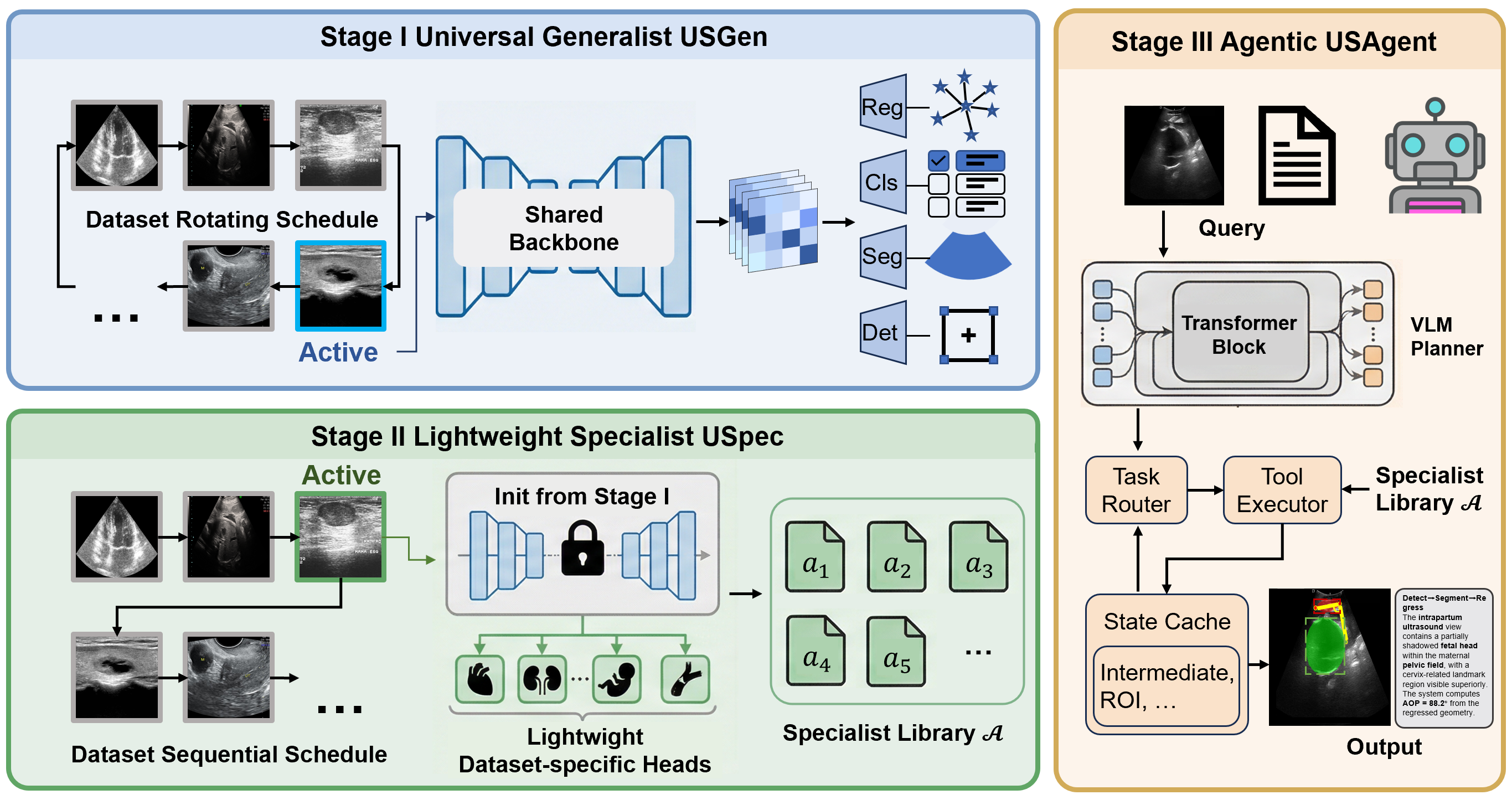}
    \caption{Overview of our proposed USTri.}
    \label{fig:overview}
\end{figure*}

As illustrated in Fig.~\ref{fig:overview}, USTri adopts a tri-stage design with increasing clinical structure. Stage I learns a shared ultrasound representation that absorbs transferable cues across organs, views, and acquisition conditions. Stage II performs lightweight dataset specialization by only finetuning compact dataset-specific heads on stage I frozen backbone, which reconciles dataset specific label spaces and improves robustness under view and device shifts. Stage III builds USAgent on top of the trained specialists, which mimics clinician workflows by selecting appropriate specialists, composing multi-step tool use, and rendering deterministic reports.

\subsection{Stage I: Universal Generalist Ultrasound Model}

In Stage I, we train USGen as a universal generalist on multi-organ, multi-view ultrasound with heterogeneous supervision. This stage targets ultrasound specific variability, including operator dependent views and acoustic artifacts, by learning transferable priors that are shared across anatomy and acquisition conditions. Importantly, USGen serves as a foundation backbone that provides stable and transferable features across organs, views, and acquisition settings, forming a consistent basis for efficient specialization and workflow composition in the subsequent stages.

Formally, for an input image $x$, a shared backbone $f_{\theta}(\cdot)$ produces a latent feature $\mathbf{z}$, and a task category head $g_{\tau}(\cdot)$ maps $\mathbf{z}$ to the prediction $\hat{y}$, where $\tau \in \{\mathrm{seg,cls,det,reg}\}$.
\begin{equation}
\mathbf{z}=f_{\theta}(x), \hat{y}=g_{\tau}(\mathbf{z})
\end{equation}

Training follows a dataset rotating schedule where we train on one dataset at a time and then switch to the next, cycling over all datasets. This schedule yields coherent optimization for each head and progressively integrates cross task commonality into the shared representation. At inference, we select the corresponding head on top of the shared backbone.

\subsection{Stage II: Lightweight Ultrasound Specialist}

In Stage II, we build USpec on top of USGen. USGen's unified training prioritizes broad transferability, so performance can be suboptimal for individual datasets under view, device, and annotation shifts. 

We freeze the USGen backbone and finetune only compact, dataset-specific heads for specialization. Compared with training a separate full model for each task, it is parameter-efficient, while improving per-dataset task alignment and robustness. For classification datasets, we further attach a lightweight adapter to recalibrate feature statistics for global decision making.

\begin{equation}
\mathbf{z}=f_{\theta}(x), \hat{y}=g_d(a_d(\mathbf{z}))
\end{equation}
where $g_d(\cdot)$ is a dataset-specific head that maps features to predictions for dataset $d$. $a_d(\cdot)$ is a lightweight adapter for classification datasets. For other datasets $a_d(\cdot)$ is the identity mapping.

Overall, Stage II produces a set of specialists that share one backbone, which improves task-wise optimality for unified ultrasound analysis while maintaining a compact and deployable representation core.

\subsection{Stage III: Agentic Ultrasound System}
To further enable end-to-end clinical ultrasound workflows with interpretable outputs, Stage III introduces USAgent, an agentic inference layer that composes the trained Stage II specialists USpec as callable tools via a biomedical VLM planner. The planner is instantiated by LLaVA-Med v1.5~\cite{li2023llava} and is restricted to a closed tool set implemented by Stage II specialists:
\begin{equation}
\mathcal{A}=\{\mathrm{Det}_d,\ \mathrm{Seg}_d,\ \mathrm{Cls}_d,\ \mathrm{Reg}_d\}_{d\in\mathcal{D}}.
\end{equation}

During inference, USAgent maintains a lightweight state that caches intermediate results, including regions of interest such as boxes or points, masks, semantic outputs such as class probabilities, and continuous measurements.

At each step, the planner selects one tool $a_k\in\mathcal{A}$  with structured parameters, and a deterministic executor runs the selected specialist and updates the state.
\begin{equation}
r_k=a_k(x,s_k,p_k),\qquad s_{k+1}=U(s_k,r_k),
\end{equation}
where $k$ is the step index, $s_k$ is the cached state at step $k$, $p_k$ are the structured tool parameters, $r_k$ is the tool output, and $U(\cdot)$ deterministically updates the state.

This enables ultrasound specific multi-step workflows such as Detect-to-Segment-to-Classify and Detect-to-Segment-to-Regress. The process terminates when the planner outputs a null action, and the final output is produced by a deterministic report renderer that aggregates cached locations, shapes, semantics, and measurements, avoiding free-form generation.

\subsection{Model Architecture}
We employ a TransUNet-style hybrid encoder ~\cite{chen2024transunet} as the shared feature extractor. Given an input image $x$, the encoder outputs token embeddings and multi-scale features through a ViT encoder, while a hybrid convolutional stem preserves an early high-resolution feature map that we treat as a dedicated feature interface for classification.

Task-specific heads are attached according to the supervision type. Segmentation follows the standard TransUNet head to output per-pixel logits. Detection uses a shallow convolutional refinement on decoded features, followed by global average pooling to regress a single normalized box. Regression is implemented as keypoint heatmap prediction~\cite{iugc, bai2026iugc}, with a small convolutional head and fixed-resolution upsampling. For classification, we do not rely on ViT tokens and instead classify from the stem feature map using adaptive average pooling and a compact MLP.

Since classification is the only non-dense task in our setting, we further add a shallow residual bottleneck stem adapter for classification datasets to recalibrate features without changing the backbone or dense decoders.

\subsection{Objective Functions}
The model is trained using a composite loss function determined by the active task in the current batch.

For segmentation, we use a multiclass Dice loss on per-pixel logits to optimize region overlap. For classification, we apply standard cross-entropy on the predicted class logits.

Regression is formulated as heatmap prediction. Given predicted heatmaps $\hat{H}$ and target heatmaps $H$, we use mean squared error:
\begin{equation}
    \mathcal{L}_{reg}=||\hat{H}-H||_2^2
\end{equation}

Point coordinates are obtained by argmax on each heatmap and converted to normalized $(x,y)$ coordinates at inference.

Detection regresses a single normalized bounding box $\hat{b}\in [0, 1]^4$ and uses an IoU-aware loss with an additional L1 term:
\begin{equation}
    \mathcal{L}_{det}=1-\mathrm{IoU}(\hat{b},b)+\alpha||\hat{b}-b||_1
\end{equation}
where $\alpha$ weights the IoU and regression penalty. $\mathrm{IoU}(\cdot,\cdot)$ is computed with a small $\epsilon$ for numerical stability, and samples with invalid box annotations are masked out during loss computation.

\section{EXPERIMENTS AND RESULTS}
\label{sec:exp}

\subsection{Datasets}

We conduct experiments on the FMC\_UIA Challenge~\cite{deng2026baseline} dataset. It is a large scale multi-center clinical ultrasound benchmark with substantial variability in acquisition devices, anatomical views, and image quality, making it suitable for evaluating generalist models under heterogeneous real world conditions.

The dataset comprises 27 subtasks spanning four task types, including segmentation, classification, detection, and regression, covering pixel wise delineation, diagnostic categorization, lesion or structure localization, and biometric measurement prediction. We follow the official benchmark protocol by training on the provided training split and reporting results on the official validation split, which is collected from unseen domains to enable a rigorous assessment of cross-domain generalization.

\subsection{Evaluation Metrics}
We follow the official challenge protocol and report task specific metrics for each subtask type.
\begin{itemize}
    \item \textbf{Segmentation:} We evaluate pixel level delineation using the Dice Similarity Coefficient (DSC) for overlap accuracy and the Hausdorff Distance (HD) for boundary fidelity.
    \item \textbf{Classification:} We assess discriminative performance with the Area Under the ROC Curve (AUC), F1 score, and Matthews Correlation Coefficient (MCC), which together provide a balanced view of ranking ability, precision recall trade off, and robustness under class imbalance.
    \item \textbf{Detection:} We measure localization quality using the Intersection over Union (IoU) between predicted and ground truth bounding boxes.
    \item \textbf{Regression:} We report the Mean Radial Error (MRE) in pixels. To ensure clinical relevance, MRE is computed on the original image resolution, rather than on the resized inputs used during training.
\end{itemize}

\subsection{Implementation Details}
We adopt a training scheme with Adam optimizer. In Stage I, we set the learning rate to $1\times10^{-4}$ for the backbone and $1\times10^{-3}$ for the task heads. In Stage II, we finetune the task decoder with a learning rate of $1\times10^{-3}$.

All images are resized to $256\times256$. For training augmentation, we apply random flips and rotations for all tasks, and additionally use random gamma and contrast jitter for segmentation, detection, and regression, with an extra random scale crop for segmentation. At inference, we use test-time augmentation constructed from the corresponding training augmentations and aggregate the predictions across augmented views.

\subsection{Results and Analysis}

\begin{table}[t]
\centering
\caption{Quantitative results on the FMC\_UIA validation set. Metrics are grouped by task type. Best results are in \textbf{bold}.}
\label{tab:results}
\begin{tabular}{clcccc}
\toprule
Task & Metric & MH-MTL & USFM & USGen & USpec \\
\midrule
\multirow{2}{*}{Seg.} & DSC$\uparrow$ & 0.7543 & 0.8862 & 0.8531 & \textbf{0.8980} \\
                    & HD$\downarrow$  & 81.18  & 31.72  & 51.08  & \textbf{27.21} \\
\midrule
\multirow{3}{*}{Cls.} & AUC$\uparrow$ & 0.9155 & 0.9337 & 0.9238 & \textbf{0.9352} \\
                    & F1$\uparrow$  & 0.7896 & 0.8417 & 0.8110 & \textbf{0.8593} \\
                    & MCC$\uparrow$ & 0.6766 & 0.7411 & 0.7013 & \textbf{0.7675} \\
\midrule
Det. & IoU$\uparrow$ & 0.2641 & 0.7909 & 0.7493 & \textbf{0.8000} \\
\midrule
Reg. & MRE$\downarrow$ & 67.43 & 21.87 & 38.30 & \textbf{18.42} \\
\bottomrule
\end{tabular}
\end{table}

Table \ref{tab:results} reports the quantitative results on the FMC\_UIA unseen-domain validation set. The official baseline MH-MTL~\cite{deng2026baseline} performs substantially worse than the foundation-style models across all tasks, indicating limited robustness under domain shift. USFM~\cite{usfm}, as a state-of-the-art self-supervised ultrasound foundation model, provides a strong improvement over the baseline and serves as a competitive reference. Our Stage I model (USGen) is already clearly stronger than MH-MTL on every metric, but still trails USFM, suggesting that generic self-supervised pretraining remains highly effective when the downstream supervision is limited to standard training. 

After the Stage II refinement, USpec consistently surpasses all methods and achieves the best results across segmentation, classification, detection, and regression. Compared with USFM, USpec improves DSC from 0.8862 to 0.8980 and reduces HD from 31.72 to 27.21, demonstrating better boundary fidelity in addition to overlap. It also brings consistent gains in classification (AUC 0.9352, F1 0.8593, MCC 0.7675), and achieves the top detection and regression performance (IoU 0.8000, MRE 18.42). Relative to USGen, USpec yields improvements on all metrics, confirming the benefit of the two-stage training strategy.

The fact that USpec outperforms USFM suggests that strong task alignment via full supervision on heterogeneous objectives can be more beneficial than generic self-supervised representations under this benchmark. USFM learns broad ultrasound features without being explicitly constrained by pixel-accurate contours, box geometry, or clinically meaningful measurement targets. In contrast, our training directly optimizes for these label-driven objectives across all task types, which likely explains the more pronounced gains on geometry sensitive metrics such as HD, IoU, and MRE. 

The gains from USGen to USpec are also expected. Stage II serves as a targeted refinement that improves robustness and precision by applying task-consistent augmentation and lightweight adaptation, making the learned representation better reflect the appearance and geometric variations encountered in unseen domains. Together, these results indicate that combining fully supervised multi-task training with a dedicated robustness-oriented refinement stage is an effective recipe for generalist ultrasound modeling under domain shift.

\begin{figure}[t]
    \centering
    \includegraphics[width=1.0\linewidth]{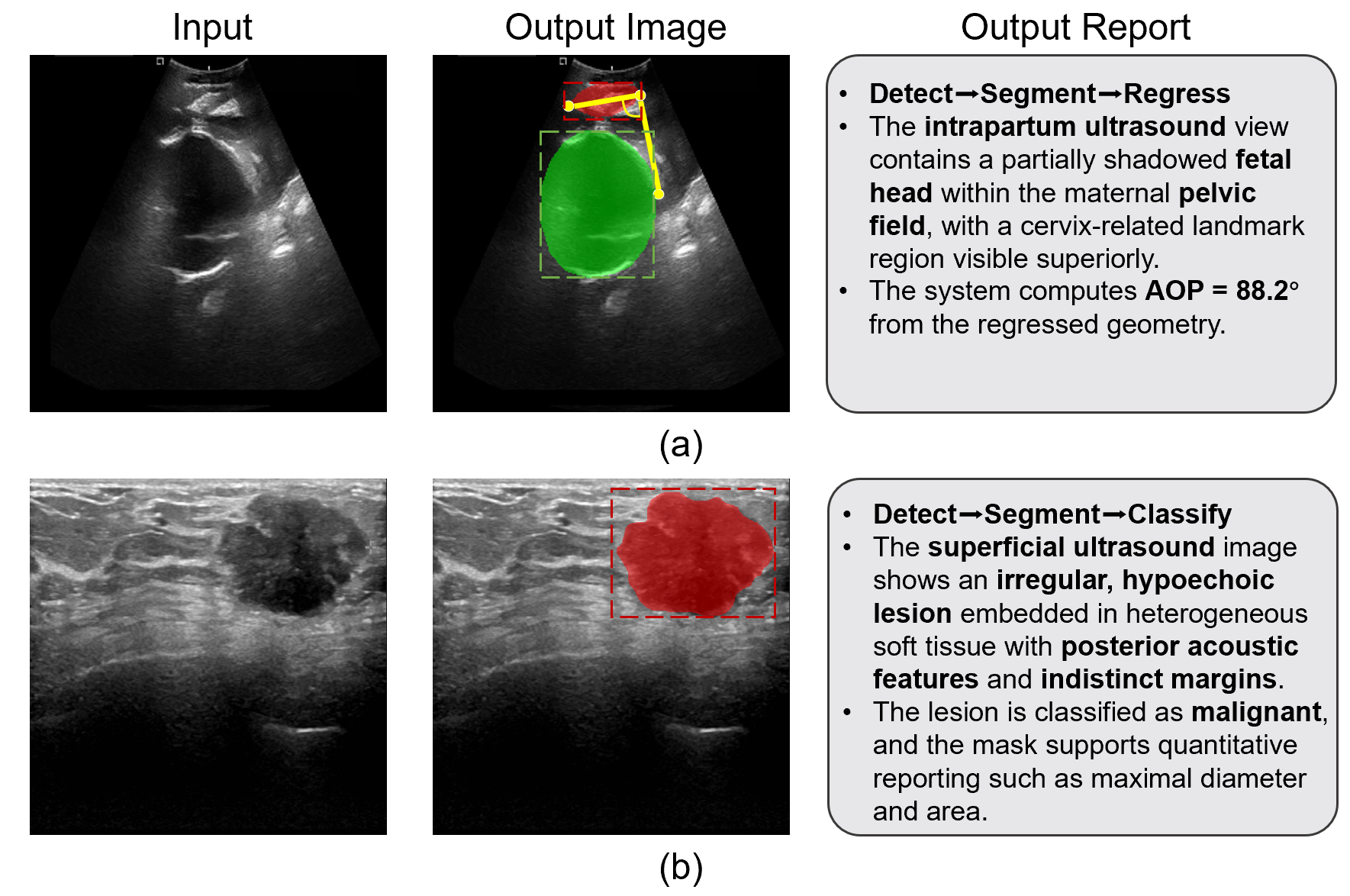}
    \caption{USAgent end-to-end workflows with verifiable evidence. (a) Intrapartum ultrasound for labor progress assessment. (b) Superficial ultrasound for breast lesion diagnosis.}
    \label{fig:2}
\end{figure}

We further provide qualitative case studies of USAgent in Fig.~\ref{fig:2}. In (a), USAgent composes a Detect, Segment, and Regress tool chain on an intrapartum ultrasound image to localize the fetal head and cervix-related landmarks, delineate the fetal-head contour, and compute the angle of progression for labor progress assessment. In (b), USAgent performs Detect, Segment, and Classify for superficial lesion assessment, producing a lesion mask and predicting a malignant diagnosis. Notably, USAgent couples each final prediction with verifiable intermediate evidence such as ROIs, masks, and measurement geometry, and renders them into a concise structured report, yielding consistent clinical-style outputs that are transparent, auditable, and readily deployable across heterogeneous ultrasound tasks.

\section{CONCLUSION}
We present USTri, a tri-stage ultrasound intelligence pipeline that evolves from a unified generalist, to parameter-efficient specialists, and finally to a clinically oriented agentic system. On the FMC\_UIA validation set, USTri achieves the best overall performance, and the agentic system further enables consistent end-to-end workflows with interpretable outputs.

\section{Acknowledgments}
This work was supported by National Key R\&D Program of China (2024YFF0507300, 2024YFF0507303), and National Natural Science Foundation of China (Grant No. 62531004).

\bibliographystyle{IEEEbib}
\bibliography{ref}

\end{document}